\pdfoutput=1

\documentclass[11pt]{article}

\usepackage[final]{acl}

\usepackage{times}
\usepackage{latexsym}

\usepackage[T1]{fontenc}

\usepackage[utf8]{inputenc}

\usepackage{microtype}

\usepackage{inconsolata}

\usepackage{graphicx}

\usepackage[textsize=tiny]{todonotes}
\usepackage{tabularx}

\usepackage{xurl}

\usepackage{listings}

\usepackage{booktabs}

\usepackage{enumitem}

\usepackage{multirow}

\title{Task-Oriented Automatic Fact-Checking with Frame-Semantics}

\author{Jacob Devasier, Rishabh Mediratta, Akshith Putta, Chengkai Li \\
  University of Texas at Arlington \\
  \texttt{cli@uta.edu} \\}

\begin{document}
\maketitle
\begin{abstract}
Automatic fact-checking is a critical tool for addressing the growing challenge of misinformation, particularly when verifying novel claims that lack previously curated evidence. Prior work has largely focused on unstructured or highly-curated data sources, limiting scalability and generalization. In this work, we introduce a novel paradigm for task-oriented automatic fact-checking using frame semantics to improve interpretability and evidence retrieval from high-volume structured datasets. We present a new pilot dataset of real-world factual claims grounded in two large-scale databases: U.S. congressional voting records and OECD country statistics. Across two case studies using these datasets, we demonstrate how frame elements can guide fine-grained retrieval and automatic fact-checking. Our experiments show that frame element-driven evidence retrieval improves recall by 14\% and 11\% over full-claim baselines in the voting and OECD case studies, respectively. We further analyze frame distributions in PolitiFact fact-checked claims and find a strong alignment with the frames targeted in our study. Overall, our results highlight frame semantics as a promising foundation for scalable, interpretable, and domain-adaptable automatic fact-checking.

\end{abstract}

\section{Introduction}

The proliferation of misinformation poses a critical challenge to the modern information ecosystem, threatening informed decision-making and public trust. Addressing this challenge requires scalable and reliable methods to verify the accuracy of claims—particularly novel ones that human fact-checkers may not have previously encountered. Automatic fact-checking has thus emerged as a crucial research area, aiming to reduce the manual burden of verifying claims by automating key steps in the fact-checking pipeline.

Much of the previous work on automatic fact-checking has focused on unstructured data, such as fact-checks from trustworthy sources, to perform claim matching~\cite{shaar-etal-2020-known} and use large language models to produce fact-check verdicts and explanations~\cite{Cheung2023FactLLaMAOI, Singhal2024EvidencebackedFC, Khaliq2024RAGARYF}. Structured tabular data from Wikipedia~\cite{Chen2019TabFactAL, Aly2021FEVEROUSFE} and scientific documents~\cite{wang-etal-2021-semeval, akhtar-etal-2022-pubhealthtab} have also been utilized to vet synthetic claims extracted from Wikipedia~\cite{bouziane-etal-2021-fabulous} and real-world claims~\cite{wang-etal-2021-semeval, akhtar-etal-2022-pubhealthtab}. 

However, these studies only consider data which have already been processed and prepared for easy consumption by readers. This reliance on highly-curated data hinders their ability to fact-check novel claims which do not have such clean evidence readily available. To bridge this gap our work proposes a novel pilot dataset of real-world factual claims related to two high-volume structured databases on U.S. congressional voting records and country statistics. These databases significantly differ from past works in both volume and detail of their tables. For example, Wikipedia tables consist of an average of 13 rows~\cite{Chen2019TabFactAL} while our database of country statistics has an average of 596,000 rows across over 400 tables. This scale of data is incompatible with most previous methods, which often rely on large language models (LLMs), as the tables themselves are too large to be analyzed due to context length limitations, even for those solutions designed for large tables~\cite{nahid-rafiei-2024-tabsqlify}.

Fact-checking novel claims requires fine-grained methodology to understand the claim and to map important parts of the claim to their relevant data. Furthermore, transparency and explainability are critical in fact-checking. We identify frame semantics~\cite{baker-etal-1998-berkeley-framenet} as a viable solution which addresses each of these points. Frame semantics is a linguistic framework that explores how meaning is encoded through structured representations called \emph{frames}. These frames capture the essential elements and relationships of the situations described by claims, supporting explainable understanding and extraction of key components from the claims. The existence of specific frames in a claim also enables fine-grained task-specific handling of the claim for fact-checking. In our paradigm, we use the frames evoked, and manually-selected frame elements, in a given claim to identify relevant database tables for automatically fact-checking it. 

Our proposed pilot dataset serves as the foundation for two case studies that demonstrate how frame semantics can be used to guide specific fact-checking processes and enhance the explainability of automatic fact-checking. In the first case study, we focus on the Vote frame~\cite{arslan-etal-2020-modeling}, which models how an \textit{Agent} (e.g., a legislator) interacts with an \textit{Issue} (e.g., a bill) through the act of voting and optionally takes a \textit{Position} (for or against) on the \textit{Issue}. These \textit{Issue} and \textit{Agent} elements are then used to implement voting-specific fact-checking methods. The second case study follows similar principles on a much more diverse set of frames applied to a high-volume, highly diverse structured database produced from hundreds of datasets collected from the Organisation for Economic Co-operation and Development (OECD). 

To evaluate these case studies, we collected 79 and 68 real-world claims for the voting records and OECD statistics, respectively. Each claim is manually annotated for its particular frames and fact-check verdict retrieved from PolitiFact. Our case studies demonstrate that frame semantics offers a valuable framework for task-specific fact-checking. Frame-semantics can also offer additional performance benefits for particular tasks. For example, on Vote and OECD-related claims, retrieving relevant evidence by querying with extracted task-specific frame elements outperformed using the entire claim by 14\% and 11\%, respectively. The performance of particular retrieval methodologies also significantly varied between the two case studies, indicating a need to select different approaches for each specific task to achieve optimal performance. Furthermore, instructing large language models (LLMs) to fact-check claims benefits from different prompts depending on the type of claim. These factors further showcase the benefit of a task-specific approach to automatic fact-checking compared to a suboptimal generalized approach. 

Finally, to understand which frames are most commonly evoked in factual claims, we surveyed claims fact-checked by PolitiFact and identified a heavy skew toward a few highly frequent frames. Notably, the frames used in our case studies are among the most frequently evoked frames in PolitiFact fact-checks. This alignment highlights the practical relevance of our approach, as it directly addresses the semantic structures most commonly encountered in professional fact-checking. 

To summarize, our contributions are as follows:
\begin{itemize}[noitemsep,topsep=2pt,leftmargin=*]
  \item We proposed a novel paradigm for task-oriented automatic fact-checking using frame-semantics.
  \item We developed a pilot dataset for automatic fact-checking of real-world claims using high-volume structured data and released it along with our source code.\footnote{\url{https://github.com/idirlab/claimlens-case-studies}} 
  \item We conducted a novel survey of frames evoked in PolitiFact fact-checks, enabling targeting high-impact frames for future studies. 
  \item We conducted two case studies on the efficacy of frame-semantics in automatic fact-checking and released a public demo.\footnote{\url{https://idir.uta.edu/claimlens/}}
\end{itemize} 

\section{Related Work}
\paragraph{Large-scale datasets.}

Previous studies~\cite{Aly2021FEVEROUSFE, gupta-etal-2020-infotabs, Chen2019TabFactAL} often use the term ``large-scale datasets'' to refer to collections with a large quantity of claims and tables. In these cases, the emphasis is on the breadth of the dataset, with a significant number of claims and corresponding tables, typically featuring a modest number of rows per table. For instance, FEVEROUS~\cite{Aly2021FEVEROUSFE} focuses on fact-checking against Wikipedia tables, which average around 13 rows per table, enabling techniques that rely on lightweight parsing and exhaustive search methods to handle the smaller, well-curated tables.

However, a key limitation of these datasets is their reliance on curated data, largely extracted from Wikipedia. Wikipedia tables, while useful for research, often represent a simplified or filtered version of underlying data from primary sources. This includes implicit decisions about which metrics to include (e.g., GDP per capita in USD vs.~PPP), how to aggregate regional data, or how to normalize column formats for readability. In contrast, our work avoids these curated intermediaries and instead operates directly on primary source data in its original, often more complex format. For example, the OECD tables used in our study contain nearly 600,000 rows per table on average.

This shift from curated to raw primary data introduces new challenges, particularly in evidence retrieval and reasoning, as systems must handle large-scale, high-dimensional tables without relying on pre-filtered or human-edited summaries. It also better reflects real-world fact-checking settings, where claims must be verified against the same raw data sources used by policymakers or researchers. 

\paragraph{Automatic fact-checking.}
Recent advances in automatic fact-checking have been largely driven by the integration of large language models (LLMs) and retrieval-augmented generation (RAG) pipelines. \citet{wang2024openfactcheck} proposed a unified framework for LLM-based systems which utilizes an internal mechanism to determine which of a selection of LLM-base models should be used to fact-check a particular claim. RAGAR~\cite{Khaliq2024RAGARYF} enhances fact-checking by using multimodal inputs and iterative reasoning. FactLLaMA~\cite{cheung2023factllama} combines a pre-trained LLaMA model with external evidence retrieval to fact-check claims. LLM-Augmenter~\cite{llmaugmenter} integrates external knowledge sources and provides feedback to improve model accuracy. \citet{Singhal2024EvidencebackedFC} mitigates misinformation generated by RAG pipelines via re-ranking retrieved documents according to a credibility score. 

For structured data, past work has studied using fine-tuning methods~\cite{zhao-etal-2022-reastap, gu-etal-2022-pasta, jiang-etal-2022-omnitab} and LLM-based methods~\cite{dater, Chen2021EvaluatingLL, Cheng2022BindingLM} for fact verification. Dater~\cite{dater} is a fact-verification system which simplifies the fact-checking process by decomposing claims into sub-questions using LLMs. It is the best-performing model on the TabFact dataset~\cite{Chen2019TabFactAL}. 
Our work does not primarily focus on fact verification; rather, our main emphasis is on evidence retrieval. Nevertheless, we employ LLM-based methods for the fact-verification step, as they offer the practical advantage of not requiring model retraining when new data becomes available.

\paragraph{Applications of frame semantics.}
\citet{madabushi2024fsragframesemanticsbased} compares search methods for finding textual entailment facts in RAG-based systems, including a frame-based method.  The authors found that their frame-based approach significantly improved recall@K compared to other search-based methods. Additionally, many previous studies have used frame semantics, or similar semantic parsing methods, in creating task-oriented systems in other NLP domains~\cite{neves-ribeiro-etal-2023-towards,gupta-etal-2018-semantic-parsing,Chen2014LeveragingFS}. 

\begin{figure*}
  \centering
  \includegraphics[width=\linewidth]{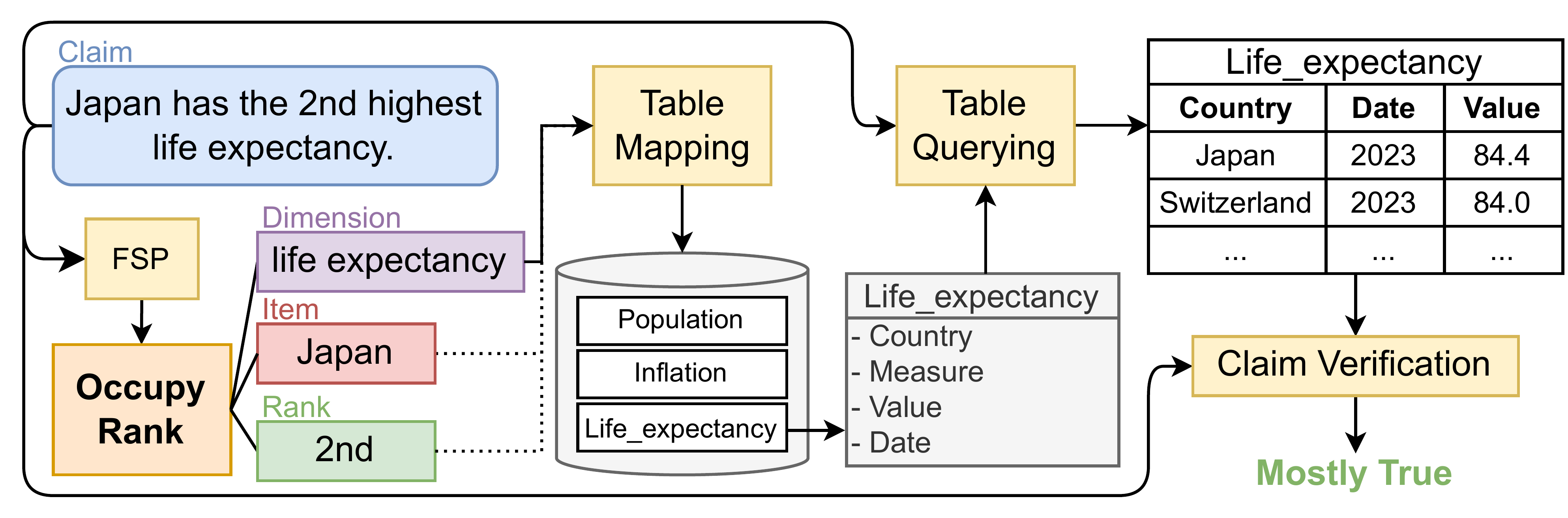}
  
  \caption{\label{fig:example}An example of our proposed paradigm. First, the frame Occupy Rank and frame elements (FEs) are extracted using a frame-semantic parser (FSP). Then, FEs are mapped to a table which can then be queried to gather evidence for the claim. Finally, the evidence and claim are passed into a fact verification model to check the truthfulness of the claim.}
\end{figure*}

\section{Fact-Checking Paradigm}
\label{sec:fc-paradigm}
We break down the task of fact-checking into three key steps: claim understanding, evidence retrieval, and fact verification. Figure~\ref{fig:example} shows an example of how our system processes the statement ``Japan has the 2nd highest life expectancy.'' 

\subsection{Claim Understanding}
In the first step, the system focuses on understanding the specifics of the claim. Frame-semantic parsing~\cite{das-etal-2014-frame} enables the extraction of frames and their frame elements from claims. 
By extracting this representation, the system gains a comprehensive understanding of the claim, which is critical for downstream tasks such as evidence retrieval and verification. For example, in Figure~\ref{fig:example}, the Occupy Rank frame is evoked, which characterizes the claim in terms of an \textit{Item} occupying a certain \textit{Rank} along a \textit{Dimension}. This semantic structure provides useful context that can support the system’s interpretation and downstream processing of the claim.

Frame-semantics not only facilitates claim understanding but also guides the fact-checking process by linking frame elements to appropriate data sources. While some frame elements, such as the \textit{Agent} and \textit{Issue} in the Vote frame, map directly to specific database tables (i.e., congress members and bills), other frames may require a predicted mapping based on the text. Figure~\ref{fig:example} illustrates an example of this predicted mapping.

\subsection{Data Collection and Evidence Retrieval}
The availability of trustworthy data is essential for fact-checking. Our system incorporates hand-selected data from existing reliable sources, such as the OECD and U.S. Congress, into an internal database. This ensures the integrity of our evidence and avoids the vulnerabilities of external sources. 

Evidence retrieval is guided by the frame elements identified in the claim understanding phase. We have manually defined the frame elements used for evidence retrieval for the frames studied in our dataset in Table~\ref{tab:frame-fes} in the Appendix. These frame elements act as filtering conditions for querying the appropriate tables or data sources. For example, in Figure~\ref{fig:example}, the \textit{Dimension} frame element is used to query the relevant table for life expectancy statistics. By focusing on specific spans of the claim (i.e., the \textit{Dimension} frame element ``life expectancy''), this method minimizes confusion and enhances retrieval accuracy.

\subsection{Fact Verification}

The final step in our pipeline involves assessing the consistency between the retrieved evidence and the original claim. Verifying the truthfulness of a claim requires not only synthesizing the retrieved evidence but also understanding how it relates to the claim and its underlying intent. For instance, in Figure~\ref{fig:example}, the claim states that Japan has the second-best life expectancy, while the data shows it actually has the highest. Although the claim is not factually precise, its core premise---that Japan has an exceptionally high life expectancy---remains valid. A robust fact verification model should ideally classify such a claim as true or mostly true. Modern LLMs are well-suited for capturing such nuances, which is why we adopt an LLM-based approach for this step.

While fine-tuned approaches have shown slightly better performance on table-based fact-checking~\cite{dater}, we employ an LLM-based fact verification approach due to their ability to handle novel inputs without needing to retrain and for their ability to provide natural language explanations of predictions. Importantly, since fact verification is not the primary focus of our work, we keep this component model-agnostic. That is, our system can accommodate any fact verification model that accepts structured evidence and a claim as input. This design choice allows for easy substitution with future state-of-the-art models as they become available.

\section{Fact-Checking Case Studies}
In this section we provide two case studies which utilize our proposed paradigm. 

\subsection{Voting Records}
PolitiFact is a leading fact-checking organization that assesses the accuracy of public statements using verifiable evidence. Voting records are a significant portion of PolitiFact's fact-checks. In this case study, we focus on automatically fact-checking voting-related claims using the Vote frame and official U.S. congressional records. Specifically, we target the \textit{Agent} and \textit{Issue} frame elements (FEs), which refer to the voting entity and the topic of the vote, respectively.

\paragraph{Datasets.}
We compiled a dataset of official U.S. congressional voting records from the Congress GitHub Repository.\footnote{\url{https://github.com/unitedstates/congress}} 
The dataset includes 342,466 bills from the 93rd to the 117th Congress. It also contains biographical information for 12,677 unique members of Congress dating back to the 1st Congress. However, comprehensive roll call voting records are only available from the 101st to the 117th Congress, totaling 7,195,798 individual votes across 22,447 roll calls. Each member of Congress cast an average of 4,230 votes during this period. Accordingly, our analysis is restricted to data from the 101st Congress onward, which reflects the full span of available voting records.

PolitiFact fact-checkers rely on evidence from official records and verified data to assess the truthfulness of claims.~\footnote{~\url{https://www.politifact.com/article/2022/mar/31/politifacts-checklist-thorough-fact-checking/}} In the context of the Vote frame, claims often reference specific congressional bills. To evaluate our system, we constructed an evaluation dataset by extracting all PolitiFact fact-checks made before April 2022 that involve claims related to the Vote frame and reference at least one congressional bill. After manual verification, we collected 79 fact-checks, along with their corresponding bills, to form the evaluation set.

\paragraph{Congress member identification.} 
To verify voting records, mapping the \textit{Agent} FE to the correct member of Congress is essential. We use database queries to match congress members whose names are similar to the words in the \textit{Agent} FE (details in Appendix~\ref{app:models}). In cases of name ambiguity we default to the more recent member. 

This stage presents several challenges. Claims often use nicknames, such as ``Sleepy Joe'' for Joe Biden or ``Meatball Ron'' for Ron DeSantis. To address this, we extracted two lists of political nicknames from Wikipedia~\cite{enwiki:1226749824, enwiki:1226728769} to map nicknames to their corresponding congress members. Similarly, some members go by shortened or preferred names, such as ``Joe Biden'' instead of ``Joseph Biden'' or ``Ted Cruz'' instead of ``Rafael Edward Cruz''. We handle this by utilizing a list~\footnote{~\url{https://github.com/carltonnorthern/nicknames}} of congress members' preferred names, supplemented by common alternatives.

\paragraph{Bill matching.}
\label{sec:bill-matching}
Identifying the correct bill based on the extracted \textit{Issue} FE is challenging because the \textit{Issue} FE can refer to various types of information, such as abstract topics (e.g., ``gun control''), specific actions or bills (e.g., the ``Inflation Reduction Act of 2022''), or outcomes of legislation (e.g., ``preventing women from getting abortions''). Additionally, bills often do not include the colloquial terms commonly used to describe them. For example, the ``STOP School Violence Act of 2018'' may be informally described as expanding access to guns in schools, even though the bill itself does not use this phrasing. Because of these challenges, keyword-based search is insufficient for accurate evidence retrieval. To address this, we employ an asymmetric semantic similarity model, as described in Section~\ref{evidence-retrieval}, enabling us to identify bills that are semantically similar to the \textit{Issue} FE, even when the style of language used in the claim and the bill differs significantly. 

\paragraph{Verifying alignment of claim to bill.} 
Determining whether a claim is refuted or supported by evidence presents several challenges. First, the system cannot rely solely on the vote (Yea or Nay) and the \textit{Position} FE (for or against).~\footnote{Note that \textit{Position} FE is extracted from claims by our frame semantic parsing model. However, only \textit{Agent} and \textit{Issue} FEs are used for evidence retrieval in voting-related claims.} Claims may be made without a \textit{Position} FE, and the relationship between the vote on a bill and the claim's \textit{Position} FE can differ. For instance, a Yea vote on a bill may not indicate support for the claimed \textit{Issue}. Consider the claim, ``DeSantis voted against allowing abortions,'' in conjunction with a Yea vote on a bill that bans abortion. Here, the vote supports the claim despite the discrepancy between the \textit{Position} FE (against) and the vote on the bill (for/yea). Second, assessing whether a claim is supported or refuted by a bill vote requires an understanding of the bill itself and its implications. To address these concerns, we instruct a large language model (LLM) with these challenges in mind using the prompt in Appendix~\ref{claim-bill-alignment-prompt}.

\paragraph{Fact verification.}
Finally, our system integrates the alignments between each bill and claim to perform fact verification over all of the retrieved evidence. We instruct a large language model (LLM) to conduct the final verification using the prompt defined in Appendix~\ref{app:fact-verification-prompts}. 

\subsection{OECD Statistics}

\paragraph{Dataset.} 
The OECD provides a wealth of trustworthy, observational data on a diverse set of statistics which serve as a strong focal point to explore a wider range of semantic frames. For example, there are statistics related to health (including healthcare coverage, health risk factors, pharmaceutical markets, etc.), financial information (including pension assets, stocks and investments, employment earnings, unemployment rates, etc.), scientific data (air pollution, greenhouse gas emissions, exposure to droughts/floods/wildfires, energy consumption, plastic leakage, etc.), and many more topics. With this dataset and the wider range of frames, we drastically expand both the scope of topics (leveraging the OECD data) and the types of claims (leveraging the frames) our system can check. 
For the OECD case study, we collected all of the data tables available on the OECD Data Explorer.\footnote{\url{https://data-explorer.oecd.org/}} We constructed an SQL database consisting of each data source, resulting in 434 tables with an average of 596,552 rows per table, totaling over 4.1 billion cells.

To evaluate our system, like the previous case study, we constructed an evaluation dataset by extracting PolitiFact fact-checks before April 2022 that cite oecd.org in their fact-check sources. After manual verification, we collected 68 fact-checks, along with their corresponding OECD data table, to form the evaluation set.

\paragraph{Relevant table identification.}
To identify the relevant table(s) to fact-check a given claim, we use a RoBERTa-based semantic similarity model~\cite{Liu2019RoBERTaAR}. To represent the table, we encode the table's name and description as text. To minimize the likelihood of missing relevant tables, we retrieve the five tables most similar to the extracted frame element (shown for each frame in Table~\ref{tab:frame-fes}) from the claim.

\paragraph{Querying retrieved tables.}
Querying the retrieved tables requires understanding of the tables, columns, and the values within them. To do this, we encode each table's name, all of its columns, and several representative sample values from each column into text. We then utilize GPT-4o to understand the requirements for verifying a claim and generate Python code to query the retrieved table(s) accordingly. We choose this approach, as opposed to standard text-to-SQL approaches~\cite{lei2024spider2} because it allows for the execution of multiple simple queries using programmed logic, rather than relying on text-to-SQL generation which can result in very complex queries for simple claims.

We represent the table schema as SQL code, following best practices from previous work~\cite{gao2023texttosqlempoweredlargelanguage}. Alongside the schema, we include representative example values from the database for each column. Our implementation is similar to \citet{nahid-rafiei-2024-tabsqlify}; however, we address critical limitations to support the high-volume OECD data. Encoded columns with fewer than five distinct values include all values, while columns with more than 100 distinct values are randomly sampled for ten representative values. For other columns we use the same RoBERTa-based semantic similarity model from the relevant table identification to select the top ten most relevant values based on the claim. Querying the databases often result in a large number of cells, so we instruct GPT-4o to refine the query by filtering out irrelevant columns, focusing on those most critical for fact-checking the claim. The prompts for this process can be found in Appendix~\ref{app:oecd-prompts}.

\paragraph{Fact verification.}
In the final step, we used an LLM-based fact verification model. We represent the extracted evidence as a list of tab-separated values using the data retrieved in the previous step. The model is instructed using the prompt in Appendix~\ref{app:fact-verification-prompts} and outputs one of five verdicts: ``false,'' ``mostly false,'' ``half-true,'' ``mostly true,'' or ``true,'' based on PolitiFact's Truth-o-Meter.~\footnote{\url{https://www.politifact.com/article/2018/feb/12/principles-truth-o-meter-politifacts-methodology-i/}}

\section{Experiments and Results}

\subsection{Datasets}
\label{sec:datasets}

\paragraph{Fact-checking frames.}
\citet{arslan-etal-2020-modeling} introduced 11 manually defined semantic frames to extend the long-running Berkeley FrameNet project~\cite{baker-etal-1998-berkeley-framenet}. Annotations for 936 sentences containing 1,029 frame-evoking targets and 3,570 frame elements are included in their dataset along with the newly-defined frames. These frames, along with the frame elements used for identifying relevant tables, can be found in Table~\ref{tab:frame-fes} in the Appendix. 

\paragraph{PolitiFact fact-checks.}
For our analysis, we utilized a dataset of 21,024 PolitiFact fact-check articles (i.e., fact-checks) collected as of April 2022. Each article contains a claim, a detailed fact-checking analysis, a verdict, and a list of sources used in the fact-check. To focus on voting-related claims, we extracted 1,552 (7.4\%) fact-checks that mention some form of ``vote.'' From this subset, we manually identified 79 fact-checks that cite congress.gov in the sources and evoke the Vote frame. For each voting claim, we manually verified the bills referenced within the fact-check are related to the claim to ensure the accuracy of our dataset. Similarly, we collected 68 fact-checks that cite oecd.org for our analysis of OECD-related claims. Each OECD claim was manually verified and mapped to the database table which can be used to fact-check it. Additional details are provided in Appendix~\ref{app:datasets}.

\begin{table}
\centering
\begin{tabularx}{\linewidth}{llcc}
\hline
\textbf{Model} & \textbf{Frames} & \textbf{Frame Acc} & \textbf{FE Acc} \\ 
\hline
\small{Random}      & Vote &            0.488 &             0.254 \\
\small{GPT-4o-mini} & Vote &            0.974 &             0.618 \\ 
\small{Vote FSP}    & Vote & \textbf{   0.990} & \textbf{   0.889} \\ 
\hline
\small{Random}          & \small{OECD} &            0.602 &             0.000 \\
\small{GPT-4o-mini}     & \small{OECD} &            0.537 &             0.372 \\ 
\small{GPT-4o-mini*}    & \small{OECD} &            0.713 &             0.461 \\ 
\small{OECD FSP}        & \small{OECD} & \textbf{   0.742} & \textbf{   0.873} \\ 
\hline
\end{tabularx}
\caption{Performance of frame-semantic parsing model on fact-checking frames compared with LLM and random choice baselines.}
\label{tab:fsp-models}
\end{table}



\begin{table}[t]
    \centering
    \resizebox{\linewidth}{!}{
    \begin{tabularx}{1.3\linewidth}{Xl}
    \hline
    \textbf{Frame} & \textbf{Samples (\%)} \\ 
    \hline
    {Taking\_sides}                                     & 7,152 (34.0\%) \\
    {Speech}                                            & 6,010 (28.6\%) \\
    {Change\_position\_on\_a\_scale}                    & 5,547 (26.4\%) \\
    {Comparing\_two\_entities}                          & 5,530 (26.3\%) \\
    {Cause\_change\_of\_position\_on\_a\_scale}         & 4,675 (22.2\%)  \\
    {Vote}                                              & 3,229 (15.4\%) \\ 
    {Comparing\_at\_two\_different\_points\_in\_time}   & 2,436 (11.6\%) \\
    {Conditional\_occurrence}                           & 2,355 (11.2\%) \\
    {Creating}                                          & 2,194 (10.4\%) \\
    {Occupy\_rank}                                      & 1,106 (5.3\%) \\
    {Oppose\_and\_support\_consistency}                 & 1,010 (4.8\%) \\
    {Recurrent\_action\_in\_Frequency}                  & 935 (4.4\%) \\
    {Ratio}                                             & 932 (4.4\%) \\
    {Capability}                                        & 869 (4.1\%) \\
    {Occupy\_rank\_via\_superlatives}                   & 767 (3.6\%) \\
    {Uniqueness\_of\_trait}                             & 497 (2.4\%) \\
    {Occupy\_rank\_via\_ordinal\_numbers}               & 329 (1.6\%) \\
    {Recurring\_action}                                 & 187 (0.9\%) \\
    {None}                                              & 12 (0.1\%) \\ 
    \hline
    \end{tabularx}}
    \caption{Distribution of semantic frame predictions in fact-checking articles using GPT-4o-mini.}
    \label{tab:frame-distribution}
\end{table}
\subsection{Frame-Semantic Parsing}
\label{sec:fsp-model}
For frame-semantic parsing, we combined the frame identification model developed by~\citet{devasier-etal-2024-robust} with a frame element identification model based on AGED~\cite{agedfsp}. By combining these state-of-the-art approaches, we unified frame and frame element identification into a single RoBERTa-based model (additional details in Appendix~\ref{app:models}). We fine-tuned two of these frame-semantic parsing models, one for each case study, using their respective frames from the training data in Section~\ref{sec:datasets}. The model used by our system is then selected based on the candidate targets in the sentence (extracted using the approach in \citealp{devasier-etal-2024-robust}). To evaluate the performance of our model for the case studies, we compared it with a baseline GPT-4o-mini model using OpenAI's structured generation. The prompt used for this model can be found in Appendix~\ref{app:fsp-model}. We evaluated each model using exact match accuracy on frame identification and argument identification (predicting the frame elements of the evoked frame). These results are presented in Table~\ref{tab:fsp-models}.

Our findings are consistent with a previous study indicating that generative LLMs tend to struggle with frame-semantic parsing~\cite{su2024unified}. Our fine-tuned frame-semantic parsing model performed much better on frame and argument identification for both case studies. A key observation was GPT-4o-mini's tendency to over-predict frame occurrence in sentences. Performance improved significantly when we constrained the analysis to only the first predicted frame (GPT-4o-mini*). Similar patterns emerged in frame element predictions, where GPT-4o-mini frequently predicted frame elements that do not exist in the frame definition or are not used in the input sentence. Examples of these are provided in Appendix~\ref{app:fsp-examples}.

\subsection{PolitiFact Survey}
To assess the broader applicability of our frame-semantics-based fact-checking system, we conducted a comprehensive coverage analysis. For this analysis, we used the frames discussed in \citet{arslan-etal-2020-modeling}, including the 11 newly defined frames from their work. Because the annotations available for the fact-checking frames are limited, we employed a zero-shot GPT-4o-mini model (detailed in Appendix~\ref{app:fsp-model}) to identify frames evoked in the PolitiFact fact-check corpus. Claims that did not trigger any of our studied frames were classified as ``None.'' The distribution of frames is presented in Table \ref{tab:frame-distribution}.

Applying GPT-4o-mini resulted in 45,772 predicted frames, indicating that it may over-predict relative to the annotated data. While the gold dataset averages 1.1 frames per claim, GPT-4o-mini predicts 2.1 frames on average. This likely stems from GPT-4o-mini assigning frames to lexical units---words that can evoke frames---that were not defined by \citet{arslan-etal-2020-modeling}. 

One indication of this can be seen with the Vote frame. While it explicitly defines only the lexical unit \textit{vote.v}, the frame could be evoked through various other expressions. For instance, the statement ``I passed a bill'' implicitly suggests a vote of affirmation. Our analysis found that only 1,198 (5.7\%) of PolitiFact fact-checks explicitly mention ``vote,'' whereas the GPT-4o-mini model predicted Vote frames at nearly three times that amount.  Although we observed numerous claims that evoked the Vote frame without explicitly using the word ``vote,'' additional research is needed to determine the true distribution of frames in factual claims.

\begin{table}
\centering
\begin{tabularx}{\linewidth}{Xlc}
\hline
\textbf{Model} & \textbf{Data} & \textbf{R@K} \\ 
\hline
\texttt{text-embedding-3-large}             & Vote  &           0.032 \\
\texttt{BM25}                               & Vote  &           0.048 \\
\texttt{distilroberta-base}                 & Vote  &           0.114 \\
\texttt{all-mpnet-base-v2}                  & Vote  &           0.115 \\
\texttt{roberta-base-v2}                    & Vote  &           0.131 \\
\texttt{multi-qa-mpnet-base}                & Vote  &           0.143 \\
\texttt{stella\_en\_400M\_v5}               & Vote  &           0.144 \\
\texttt{stella\_en\_1.5B\_v5}               & Vote  &           0.144 \\
\texttt{distilbert-tas-b}                   & Vote  & \textbf{  0.165} \\

\hline
\texttt{stella\_en\_1.5B\_v5}            & \small{OECD}   &          0.126 \\ 
\texttt{BM25}                            & \small{OECD}   &          0.323 \\ 
\texttt{distilbert-tas-b}                & \small{OECD}   &          0.505 \\ 
\texttt{text-embedding-3-large}          & \small{OECD}   &          0.610 \\ 
\texttt{roberta-base-v3}                 & \small{OECD}   &          0.642 \\ 
\texttt{roberta-base-v2}                 & \small{OECD}   & \textbf{ 0.726} \\ 
\hline
\end{tabularx}

\caption{Performance comparison of different semantic similarity models on extracted frame elements. K=10 for Vote and K=5 for {\small OECD}.}
\label{tab:similarity-model}
\end{table}

\subsection{Evidence Retrieval}
\label{evidence-retrieval}
\paragraph{Similarity model selection.}
To retrieve relevant information for fact-checking, we first evaluated different semantic similarity models. We experimented with several models from the MTEB leaderboard.~\footnote{\url{https://huggingface.co/spaces/mteb/leaderboard}}  Table~\ref{tab:similarity-model} presents the best-performing models, including Stella~\cite{zhang2025jasperstelladistillationsota}, RoBERTa~\cite{Liu2019RoBERTaAR}, DistilBERT-TAS-B~\cite{tasb}, OpenAI's text-embedding-3,~\footnote{~\url{https://platform.openai.com/docs/models}} and BM25.~\footnote{~\url{https://github.com/dorianbrown/rank_bm25}}

Our analysis revealed that different models excel in specific verification contexts. RoBERTa demonstrated superior performance in identifying relevant OECD tables, while DistilBERT-TAS-B achieved the best results in matching claims to bills. These findings suggest that optimal fact-checking systems should employ task-specific embedding models rather than a one-size-fits-all solution.

\begin{table}
\centering
\begin{tabularx}{\linewidth}{Xllc}
\hline
\textbf{Model} & \textbf{Query} & \textbf{Data} & \textbf{R@K} \\ 
\hline
\texttt{\small distilbert-tas-b}   & Full claim        & Vote      &           0.143 \\ 
\texttt{\small distilbert-tas-b}   & Issue FE          & Vote      & \textbf{  0.165} \\ 
\texttt{Max Possible}                & -                 & Vote      & \textit{  0.568} \\

\hline
\texttt{RoBERTa (v2)} & Full claim        & \small{OECD}      &           0.653 \\ 
\texttt{RoBERTa (v2)} & FE                & \small{OECD}      & \textbf{  0.726} \\ 
\texttt{Max Possible}        & -                 & \small{OECD}      & \textit{  0.910} \\
\hline
\end{tabularx}
\caption{Performance of different query representations in each case study. Evaluations are performed on table matching (OECD) and bill matching (Vote), and use Recall@5/10, respectively.}
\label{tab:query-method}
\end{table}

\paragraph{Query representation.}

We explored two different methods for constructing retrieval queries: using the complete claim text and using only the frame elements (FEs) extracted based on the evoked frame (detailed in Appendix~\ref{app:query-fes}). For each frame, we selected specific frame elements (Table~\ref{tab:frame-fes}) which are most relevant for evidence retrieval. We evaluated these query representations using the best-performing similarity models identified in Table~\ref{tab:similarity-model}.

Table~\ref{tab:query-method} presents the recall@K metrics for each representation approach, with K=5 for OECD claims and K=10 for Vote claims. We also established baseline metrics for evidence availability (Max Possible) in our database. For voting-related claims, we calculated the percentage of referenced bills present in our database, while for OECD claims, we assessed the availability of relevant datasets.

Our results demonstrate that queries constructed from frame elements consistently outperformed full-claim queries across both domains. This finding aligns with research by \citet{madabushi2024fsragframesemanticsbased} showing that frame-semantics-based search strategies enhance retrieval effectiveness.

The performance varied significantly between domains. For voting-related claims, even our best-performing model (DistilBERT-TAS-B) achieved a modest recall@10 of 0.165, reflecting the complexity of bill matching. In contrast, OECD claim verification was easier for the model, with RoBERTa achieving a recall@5 of 0.726, likely due to the more constrained search space. Congress member identification performed very well with an exact match accuracy of 97.5\%.

\paragraph{Query generation.}
The effectiveness of generated queries varied significantly between the two domains: 62\% of OECD claim queries successfully retrieved relevant data, compared to only 36\% for voting claims. These retrieval rates effectively establish upper bounds for our fact verification capabilities, as meaningful fact-checking requires successful evidence retrieval. 

\begin{table}
\centering
\begin{tabularx}{\linewidth}{Xlc}
\hline
\textbf{Model} & \textbf{Dataset} & \textbf{Accuracy} \\ 
\hline
\small{GPT-4o Naive}  & Vote        & 0.044 \\ 
\small{Ours w/ Irrelevant}           & Vote  & 0.076 \\ 
\small{Ours w/o Irrelevant}          & Vote  & \textbf{0.207} \\ 
\hline
\small{GPT-4o Naive}                      & \small{OECD} & 0.073 \\ 
\small{Ours w/ Irrelevant}          & \small{OECD} & 0.214 \\
\small{Ours w/o Irrelevant}          & \small{OECD} & \textbf{0.429} \\
\hline
\end{tabularx}
\caption{End-to-end fact verification performance on the two case studies. The best-performing models are bolded.}
\label{tab:verification}
\vspace{-0.2cm}
\end{table}

\subsection{Fact Verification}

We evaluated our fact verification system by comparing its predicted verdicts against the ground truth verdicts from PolitiFact fact-checks (Table~\ref{tab:verification}). For consistency in our analysis, we consolidated PolitiFact's ``pants on fire'' rating with ``false''.

To assess whether our system genuinely relies on retrieved evidence rather than pre-existing knowledge, we established a baseline using GPT-4o without access to any external data (GPT-4o Naive). Our analysis encompassed claims across both voting records and OECD datasets, with performance measured against human-annotated verdicts.

The baseline GPT-4o Naive model performed poorly on both case studies, suggesting that the model's internal knowledge alone is insufficient for accurate fact verification. When we excluded cases where our system failed to retrieve relevant evidence (w/o Irrelevant), we observed substantial improvements in accuracy: an increase of 13.1 percentage points for voting-related claims and 21.5 percentage points for OECD-related claims. This improvement demonstrates the critical role of successful evidence retrieval in fact verification accuracy. Detailed examples of the system's reasoning and explanations can be found in Appendix~\ref{app:vote-alignment-examples}.

Our analysis revealed a bias in the model's predictions toward classifying claims as false. This bias likely stems from the nature of contradiction-based verification where disproving a claim requires finding just one contradictory piece of evidence, while proving a claim often necessitates comprehensive evidence supporting all aspects of the assertion. This suggests potential areas for future improvement in balancing verification criteria.

\section{Conclusion}
In this work, we introduced a novel paradigm for task-oriented automatic fact-checking by leveraging frame semantics to enhance structured understanding and evidence retrieval. Through the development of a pilot dataset composed of real-world claims and two focused case studies, we demonstrated how frame-semantic parsing can guide the fact-checking process with structured data, improving retrieval performance and explainability. Our findings suggest that semantic frames offer a meaningful way to bridge the gap between complex claims and high-volume structured datasets.

The empirical results highlight the advantages of using frame elements for evidence retrieval, with improvements in recall across both voting-related and OECD-based fact-checking tasks. By aligning claims with relevant structured data, our approach not only enhances accuracy but also provides a more interpretable verification process. Furthermore, our survey of fact-checking frames reveals promising avenues for future research, particularly in expanding the coverage of high-impact frames and refining claim-evidence alignment techniques.

While our system is limited by the availability of trustworthy data, our findings highlight a critical research gap left by approaches that focus on smaller tabular data. This work showcases the use of high-volume structured databases for automatically verifying novel claims. Expanding beyond our pilot datasets to large-scale collections of claims and corresponding evidence from structured sources remains an important next step.

\section*{Limitations}

Despite the advancements made in this study, our system's limitations should be acknowledged. While this study focuses primarily on improving evidence retrieval, it does not comprehensively evaluate different fact verification approaches. Fact verification is critical for automatic fact-checking, and a deeper investigation into various verification strategies, including fine-tuned models and LLM-based approaches, could further enhance system performance.

Another limitation is that our system’s effectiveness is constrained by the availability of reliable, manually curated data. For claims that lack relevant data in our databases (i.e., OECD tables or U.S. congressional records), the system cannot retrieve evidence, which limits its fact-checking capabilities. This limitation demonstrates the importance of expanding data sources and improving coverage in future iterations of the system.

Finally, the method relies on existing FrameNet frames. FrameNet versions for languages other than English are often incomplete or entirely unavailable. As a result, the current method may not be effective for fact-checking in non-English languages.

\section*{Ethics and Risks}

Automated fact-checking systems, such as the one presented in this work, bring both opportunities and ethical challenges. One key concern is the potential spread of misinformation due to model limitations or errors. Users may over-rely on AI verdicts, leading to the amplification of false positives or negatives, especially in politically sensitive contexts. Biases present in both the models (e.g., GPT-4o-mini, RoBERTa) and the datasets (e.g., PolitiFact, OECD data) could skew fact-checking outcomes, favoring certain political narratives or ideologies.

Another concern is the transparency and accountability of AI systems. If users are unaware of how models arrive at their conclusions, it may be difficult to hold these systems accountable for erroneous outcomes. This opacity could diminish trust in both the fact-checking tool and the institutions that deploy it. Moreover, the collection and processing of political data raise potential privacy concerns, especially regarding the use of public records in ways individuals may not expect.

Marginalized communities may also be disproportionately affected by such systems, as they might misinterpret claims relevant to those groups or lack sufficient representation in training data. Similarly, the system could be exploited by adversaries who craft ambiguous or misleading claims designed to confuse AI, leading to manipulated fact-checks.

Lastly, the over-dependence on specific sources such as PolitiFact or congress.gov could limit the tool's scope. Ethical development of such systems requires attention to these risks to ensure fairness, accuracy, and social responsibility in their use.

\section*{Acknowledgments}
This work is partially supported by the National Science Foundation under Grant 2346261. We also extend our gratitude to the Texas Advanced Computing Center (TACC) for providing compute resources used in this work's experimentation.

\bibliography{anthology,custom}

\appendix

\section{Data Quality}
\label{app:datasets}

\begin{table*}
\centering
\begin{tabularx}{\textwidth}{llX}
\hline
\textbf{Frame} & \textbf{Frame Element} & \textbf{Frame Definition} \\ 
\hline
\small{Occupy\_rank} & \small{Dimension} & \small{Items occupying a certain Rank within a hierarchy.}\\ 
\small{Occupy\_rank\_via\_superlatives} & \small{Dimension} & \small{An Item occupying a Rank specified by a superlative.}\\ 
\small{Comparing\_two\_entities} & \small{Comparison\_criterion} & \small{Comparing two entities using a Comparison\_criterion and qualifying with a Degree.} \\ 
\small{Comparing\_at\_two\_different\_ points\_in\_time} & \small{Comparison\_criterion} & \small{Comparing an Entity with itself at two different points in time using a Comparison\_criterion and qualifying with a Degree.} \\ 
\small{Cause\_change\_of\_position\_ on\_a\_scale} & \small{Item} & \small{Words that indicate an Agent or Cause affects the position of an Item on a scale.} \\ 
\small{Capability} & \small{Event} & \small{An Entity meets the pre-conditions for participating in an Event.} \\ 
\small{Vote} & \small{Agent, Issue}  & \small{An Agent makes a voting decision on an Issue.} \\ 
\hline
\end{tabularx}
\caption{Frames used for OECD case study, along with the frame elements extracted from corresponding frames.}
\label{tab:frame-fes}
\end{table*}

\paragraph{Annotators.}
The OECD and voting-related claims were annotated by three students, including one senior PhD student with expertise in frame-semantic parsing. The PhD student provided the final ground truth annotations. The two other students independently provided their annotations, and if they matched the PhD student's annotations, they were kept. No disagreements occurred during annotation.

\begin{table}
    \centering
    \begin{tabularx}{\linewidth}{Xc}
        \textbf{Model} & \textbf{Parameters} \\ 
        \hline
        \href{https://huggingface.co/sentence-transformers/msmarco-distilbert-base-tas-b}{msmarco-distilbert-base-tas-b} & 66M \\ 
        \href{https://huggingface.co/sentence-transformers/msmarco-roberta-base-v2}{msmarco-roberta-base-v2} & 125M \\ 
        \href{https://huggingface.co/sentence-transformers/msmarco-roberta-base-v3}{msmarco-roberta-base-v3} & 125M \\ 
        \href{https://huggingface.co/sentence-transformers/stella_en_1.5B_v5}{stella\_en\_1.5B\_v5} & 1.5B \\ 
        \href{https://openai.com/index/new-embedding-models-and-api-updates/}{text-embedding-3-large} & unknown \\ 
        \hline
    \end{tabularx}
    \caption{\label{tab:model_reproducibility}Models and their parameter sizes.}
\end{table}

\paragraph{Voting claims.}
Our dataset of voting-related claims consists of original claims collected from PolitiFact, corresponding URLs, manually cleaned claim text, and references to legislative bills cited in the sources. The dataset contains 79 claims referencing 190 bills, averaging 2.4 bills per claim. However, this distribution is skewed, with a median of only 1 bill per claim due to a few claims referencing a large number of bills.

Fact-checking activity was concentrated in election years, with 86\% of the 79 claims fact-checked in 2016, 2018, or 2020. Health insurance was the most frequently mentioned topic, appearing in 11 claims. The terms ``for'' and ``against'' were used in 28 and 27 claims, respectively, suggesting a frequent focus on arguments in favor of or opposition to specific policies.

No single individual was disproportionately fact-checked among the 79 claims; Donald Trump had the most fact-checked claims (3), followed by Hillary Clinton (2). Claim lengths ranged from 7 to 62 words, with an average of 18.4 and a median of 17 words. 

\paragraph{OECD claims.}
The OECD-related claims were more evenly distributed across years, with the top three years (2016, 2014, 2011) accounting for 46\% of the 68 claims. The United States was the most frequently mentioned country, appearing explicitly in 25\% of the claims. Given PolitiFact's U.S. focus, implicit references to the U.S. are likely higher. 

Taxes were the most common topic, with approximately 20\% of the claims mentioning tax-related topics. Claim lengths ranged from 9 to 41 words, with an average of 17.7 and a median of 15 words.

Almost all OECD claims contained a single frame, with only one instance containing multiple frames (Occupy\_rank and Comparing\_two\_entities). The most frequent frames were Comparing\_two\_entities (25 claims), Occupy\_rank\_via\_superlatives (12 claims), and Occupy\_rank (8 claims), indicating a strong emphasis on ranking and comparison within these claims.

\begin{table*}[t]
\centering
\small
\begin{tabular*}{\textwidth}{@{\extracolsep{\fill}} p{0.25\textwidth} p{0.25\textwidth} p{0.08\textwidth} p{0.14\textwidth} p{0.14\textwidth} @{}}
\toprule
\textbf{Claim} & \textbf{Bill Title} & \textbf{Vote} & \textbf{Alignment} & \textbf{Prediction} \\ 
\midrule
Marco Rubio voted against the bipartisan Violence Against Women Act & 112 S 1925 Violence Against Women Reauthorization Act of 2012 & Nay & Supports & True \\ 
\addlinespace[0.5em]
& 17 HR 3233 National Commission to Investigate the January 6 Attack on the United States Capitol Complex Act & Nay & Irrelevant &  \\  
\addlinespace[0.5em]
& 117 HR 350 Domestic Terrorism Prevention Act of 2022 & Nay & Irrelevant &  \\ 
\midrule
Chuck Grassley was voting to slash Medicare when voting against the debt ceiling bill & 117 S 610 Protecting Medicare and American Farmers from Sequester Cuts Act & Nay & Supports & Mostly False \\
\addlinespace[0.5em]
& 117 S 1301 Promoting Physical Activity for Americans Act & Nay & Irrelevant &  \\
\addlinespace[0.5em]
& 17 HR 1868 To prevent across-the-board direct spending cuts, and for other purposes & Yea & Refutes &  \\
\bottomrule
\end{tabular*}
\caption{Examples of voting-related claims with the corresponding retrieved bills, votes on the bills, vote-claim alignment, and fact verification prediction.}
\label{tab:vote-claim-alignment}
\end{table*}

\section{Studied Frames}
\label{app:query-fes}
Table~\ref{tab:frame-fes} provides a comprehensive list of frames studied in this work along with the frame elements used to predict table mappings. The selected frames were chosen based on their relevance to factual claims in the domains of voting and country statistics. Our analysis primarily focused on frames that involve ranking, comparisons, and legislative actions, as these are commonly invoked in fact-checking scenarios.

\section{Reproducibility}
Ensuring reproducibility is a core principle of this work. We provide details about our experimental setup, computational resources, software dependencies, and models used.

\subsection{Computational Resources}
We trained our frame-semantic parser on an NVIDIA GTX 1080 Ti GPU. A single training run took approximately 6 hours. No hyperparameter tuning was performed, and the model was trained using default settings unless explicitly stated. The OECD database requires roughly 85GB of storage and the voting records requires 1GB of storage.

\subsection{Software and Dependencies}
Our implementation is based on PyTorch and Hugging Face’s ``transformers'' and ``sentence-transformers'' libraries. A full list of dependencies and their versions is provided in our public repository. GitHub Copilot was used in the development of some of our code to assist with boilerplate functions and standard NLP preprocessing tasks.

\subsection{Software License}
All source code and dataset annotations are released under the MIT License.

\subsection{Models}
\label{app:models}
Unless stated otherwise, all models used in our experiments are the ``base'' variant. All semantic similarity models are sourced from the   ``sentence-transformers'' library.\footnote{\url{https://huggingface.co/sentence-transformers}} Table~\ref{tab:model_reproducibility} provides an overview of the models used and their parameter count.

\paragraph{Frame-semantic parsing.}
Our frame-semantic parsing model builds on AGED, which achieves state-of-the-art performance in identifying the spans of frame elements for a given frame in a sentence. While AGED focuses on span detection, \citet{devasier-etal-2024-robust} propose a complementary training strategy that improves frame identification by incorporating additional negative examples during training. Both models use similar underlying architectures, which allowed us to integrate their strengths. Specifically, we augmented AGED with an additional classification layer to predict the correct frame and adopted the robust training procedure introduced by Devasier et al. For our base encoder, we used RoBERTa instead of BERT, as it provides stronger contextual representations. The resulting model demonstrated strong empirical performance, so we retained this architecture for all downstream tasks.

\paragraph{Voting agent matching.}
Our agent matching function retrieves the BioGuide ID~\footnote{\url{https://www.congress.gov/help/field-values/member-bioguide-ids}} of a political figure from the voting records database. It extracts the agent's name using frame element spans and checks if it is a known political nickname, replacing it with the full name if applicable. If not, it uses a SpaCy NER model\footnote{\url{https://spacy.io/api/entityrecognizer}} to extract PERSON named entities. The function first attempts an exact match then a fuzzy match using SQL's LIKE operator if the exact match fails. If neither succeeds, it constructs a multi-condition query, incorporating nickname expansions where applicable.

\section{Examples}
To provide a clearer understanding of our system’s outputs, we include example cases from both the voting and OECD datasets. These examples illustrate how our models process claims, align them with retrieved documents, and generate predictions.  

\begin{table*}
    \centering
    \begin{tabularx}{\linewidth}{p{0.2\textwidth}p{0.1\textwidth}p{0.45\textwidth}p{0.2\textwidth}}
        \hline
        \textbf{Claim} & \textbf{Predicted Verdict} & \textbf{Explanation} & \textbf{Fact-Check} \\ \hline

        Americans work way more than an average of industrialized countries around the world. & Mostly True & \small{The data shows that the United States has an annual average of 1829 hours worked per person, which is higher than many other industrialized countries in the dataset, such as Germany (1400.4 hours), France (1518.0 hours), and Japan (1729.0 hours). However, there are some countries like Mexico (2229.0 hours), Korea (2076.0 hours), and Chile (1994.0 hours) with higher working hours than the U.S. This indicates that while Americans work more than many other industrialized countries, they do not work more than all. The claim is mostly true but lacks the nuance that there are industrialized countries where people work more hours.} & \href{https://www.politifact.com/factchecks/2014/feb/09/keith-ellison/keith-ellison-says-americans-work-way-more-others/}{PolitiFact} \\ 
        
        The United States is the ONLY industrialized country without universal healthcare. & True & \small{The retrieved data indicates that countries such as Italy, the Netherlands, Spain, Canada, Korea, Australia, France, Germany, the United Kingdom, and Sweden all have near-universal health insurance coverage through government or compulsory schemes, with coverage rates often very close to 100\% of the population. In contrast, the United States has only 38.1\% of its population covered by government or compulsory health insurance, suggesting that it lacks universal healthcare coverage. This supports the claim that the United States is the only industrialized nation without such a system based on the data provided.} & \href{https://www.politifact.com/factchecks/2019/jun/21/mark-pocan/universal-health-care-diagnosis-mark/}{PolitiFact} \\ 

        \hline
        
    \end{tabularx}
    \caption{\label{tab:oecd_fact_check_explanations}OECD fact-check example predictions and  explanations.}
\end{table*}

\subsection{Vote-Claim Alignment}
\label{app:vote-alignment-examples}

Table~\ref{tab:vote-claim-alignment} presents two sample claims from the voting dataset along with their top three retrieved bills. For each retrieved bill, we include the computed vote-claim alignment classification (one of Supports, Refutes, Irrelevant). This alignment process plays a crucial role in determining the factual accuracy of voting-related claims.

\subsection{Fact Verification Predictions}
\label{app:fsp-examples}

\paragraph{Voting claims.}
For voting-related claims, we provide two examples where our model predicts the veracity of claims based on retrieved bills. Table~\ref{tab:vote-claim-alignment} includes the claims, retrieved evidence, and the final prediction class.

\paragraph{OECD claims.}
For OECD-related claims, we provide two sample predictions and their corresponding explanations in Table~\ref{tab:oecd_fact_check_explanations}. These examples demonstrate how our model interprets numerical comparisons, tax-related claims, and ranking statements to verify factual assertions.

\section{LLM Prompts}

\subsection{Frame-Semantic Parsing}
\label{app:fsp-model}
\lstset{
    basicstyle=\small\ttfamily,
    backgroundcolor=\color{gray!10}, 
    linewidth=\columnwidth,
    breaklines=true,
    frame=single, 
    rulecolor=\color{black}, 
    showstringspaces=false, 
    numbers=left, 
    numberstyle=\tiny\color{gray}, 
    xleftmargin=0em, 
    numbers=none, 
    moredelim=**[is][\color{gray!70}]{@}{@} 
}
\begin{lstlisting}[caption={Frame-semantic parser for the Vote frame.}, label={listing:vote-fsp}]
Identify if a 'Vote' semantic frame is evoked in a given sentence. If it is, extract and list the relevant frame elements associated with the voting event.

A 'Vote' frame is defined as: An Agent makes a voting decision on an Issue.

Frame elements to identify:
- Agent: The conscious entity (usually a person) executing the voting decision.
- Issue: The subject or matter that the Agent is voting on with a particular position.
... [omitted for brevity]

# Notes

- If a frame element is not mentioned in the sentence, use "" for its value.
- Carefully distinguish between elements; some may have overlapping characteristics.
- Consider synonyms or variations of terms related to voting when analyzing the sentence.
- Frame elements should quote exactly from the input.
\end{lstlisting}

\lstset{
    basicstyle=\small\ttfamily,
    backgroundcolor=\color{gray!10}, 
    linewidth=\columnwidth,
    breaklines=true,
    frame=single, 
    rulecolor=\color{black}, 
    showstringspaces=false, 
    numbers=left, 
    numberstyle=\tiny\color{gray}, 
    xleftmargin=0em, 
    numbers=none, 
    moredelim=**[is][\color{gray!70}]{@}{@} 
}
\begin{lstlisting}[caption={Frame-semantic parser for OECD frames.}, label={listing:oecd-fsp}]
Identify semantic frames evoked by a given factual claim and extract relevant frame elements for each identified frame.

Consider the predefined semantic frames and their elements:

- Occupy_rank_via_superlatives:
  - Item: Entity occupying the rank.
  - Rank: Rank held, often defined by a superlative.
  ... [omitted for brevity]

- Occupy_rank:
  - Item: Entity occupying the rank.
  - Rank: Rank held.
  ... [omitted for brevity]

- Change_position_on_a_scale:
 - Item: Entity whose position on a scale changes.
 - Attribute: Property or scale of the change.
  ... [omitted for brevity]

- Comparing_two_entities:
  - Entity_1: First entity in comparison.
  - Entity_2: Entity serving as the comparison point.
  ... [omitted for brevity]

- Comparing_at_two_different_ points_in_time:
  - Entity: Entity compared to itself at different times.
  - First_point_in_time: First time period in comparison.
  ... [omitted for brevity]

# Steps

1. Carefully read and analyze the factual claim.
2. Determine which semantic frame(s) are invoked by the claim.
3. For each evoked frame, extract the appropriate frame elements directly from the claim.
4. Document each evoked frame and the extracted elements.

# Output Format

- List each evoked frame followed by its extracted elements in a structured form.
- Ensure all extracted frame elements are clearly labeled and match the exact inputs.

# Notes
- Some claims may trigger multiple frames; ensure all applicable frames are evaluated.
- If specific frame elements cannot be determined, note their absence for completeness.
- If a frame element does not explicitly appear in the claim, leave it blank.
\end{lstlisting}

\subsection{Claim-Bill Alignment}
\label{claim-bill-alignment-prompt}
\lstset{
    basicstyle=\small\ttfamily,
    backgroundcolor=\color{gray!10}, 
    linewidth=\columnwidth,
    breaklines=true,
    frame=single, 
    rulecolor=\color{black}, 
    showstringspaces=false, 
    numbers=left, 
    numberstyle=\tiny\color{gray}, 
    xleftmargin=0em, 
    numbers=none, 
    moredelim=**[is][\color{gray!70}]{@}{@} 
}
\begin{lstlisting}[linewidth=\columnwidth,breaklines=true,caption={Claim-Bill Alignment}, label={listing:claim_bill_alignment}]
Given the following factual claim, bill summary, and vote on the bill, evaluate whether the content of the bill summary and the voting record align with the given claim. You may consider factors such as the main objectives of the bill and unintended or implicit consequences. Your task is to determine if the information provided in the bill summary and the voting record supports or refutes the given factual claim. Return your explanation and one of the following labels in JSON format.

Bill Summary:
{summary}

Vote: {vote_type}
Claim: {claim}

Labels:
Supports - The vote on this bill directly or indirectly supports the claim.
Refutes - The vote on this bill explicitly refutes the claim.
Inconclusive - The vote on this bill does not provide enough information to support or refute the claim.
Irrelevant - The vote on this bill is not relevant to the claim at all.
\end{lstlisting}

\subsection{OECD Data Query}
\label{app:oecd-prompts}

\lstset{
    basicstyle=\small\ttfamily,
    backgroundcolor=\color{gray!10}, 
    linewidth=\columnwidth,
    breaklines=true,
    frame=single, 
    rulecolor=\color{black}, 
    showstringspaces=false, 
    numbers=left, 
    numberstyle=\tiny\color{gray}, 
    xleftmargin=0em, 
    numbers=none, 
    moredelim=**[is][\color{gray!70}]{@}{@} 
}
\begin{lstlisting}[caption={Retrieve Data for Fact-Checking Claim}, label={listing:retrieve_data}]
Your task is to write a Python function named retrieve_data() that retrieves data for fact-checking the claim:
{claim}

Given the following database schemas from OECD_Data.db:
{table_descriptions}

Instructions:
- Use LIKE only for textual columns. For numerical columns (e.g., year, value), use appropriate comparison operators like = or <=.
- Do not modify the database.
- The function must contain all necessary imports inside it and take no parameters, though passing parameters (e.g., file paths or claim details) should be considered in future iterations.
- Return the data in pandas DataFrames. Multiple DataFrames can be returned as a list if needed.
- If no relevant data is found, return 'Data is not Available'.
- Use only actual values found in the columns. If aggregation is possible (e.g., summing or averaging categories), return aggregated results, unless the claim specifies otherwise. Use appropriate aggregation methods based on the data.
- Always return relevant columns (avoid '*' when possible). Select the necessary columns based on the claim.
- Treat 'we' in the claim as referring to 'United States', but consider the context of the claim for potential exceptions.
- If any 'unit_of_measure' is 'National currency', use the 'USD_value' column instead of 'value'.
- Use the nearest available date based on the provided dictionary: {nearest_dates}. If no close date is found, return 'Data is not Available'.
- For multiple tables, create multiple queries as needed. If combining data from multiple tables is required, ensure consistency in merging and handling differences in structure.
- Do not filter by 'country' or 'unit_of_measure'; that will be handled later.

Output:
- Return a list of pandas DataFrames or 'Data is not Available' if no relevant data is found.
\end{lstlisting}

\lstset{
    basicstyle=\small\ttfamily,
    backgroundcolor=\color{gray!10}, 
    linewidth=\columnwidth,
    breaklines=true,
    frame=single, 
    rulecolor=\color{black}, 
    showstringspaces=false, 
    numbers=left, 
    numberstyle=\tiny\color{gray}, 
    xleftmargin=0em, 
    numbers=none, 
    moredelim=**[is][\color{gray!70}]{@}{@} 
}
\begin{lstlisting}[linewidth=\columnwidth,breaklines=true,caption={Clean Data for Fact-Checking Claim}, label={listing:clean_data}]
Your task is to write a Python code function named `clean_data()` to filter and clean data for fact-checking the claim {claim} from `OECD_Data.db`.  

Extracted dataframes from the previous step:  
{list_of_dfs}  

These dataframes were generated using this code:  
{code}  

The schema for these dataframes is:  
{schema_str}  

Notes:  
- The function must have all imports inside it and take no parameters.  
- This function will run independently of the given code, meaning you need to re-extract the data from the database to clean it for fact-checking the claim:  
  {claim}  
- Do not modify the original way the data was extracted unless the input code is very incorrect; just add more filters/conditions based on the claim.  
- Keep the same aggregated structure from the original extraction code unless the claim specifies otherwise.  
- Filter by the nearest date: {nearest_date}.  
- Exclude tables only if they are irrelevant to the claim.  
- Use only the actual tables and data that were extracted; do not make up new tables.  
- Add more filters to narrow down to claim-relevant metrics; try to have one value per non-numeric column wherever possible.  
- Claim-relevant metrics may appear in columns other than 'measure', so reference the claim to identify the appropriate columns to filter by.  
- Use the retrieval code to identify useful information for your filters, if needed.  
- When filtering by 'unit_of_measure', ensure to use a standardized metric (e.g., a common currency) to simplify further analysis.  

Output:  
- A list containing cleaned pandas DataFrames based on the claim. If the code is run and the data is not available, return 'Data is not Available'.
\end{lstlisting}

\subsection{Fact Verification}
\label{app:fact-verification-prompts}

\lstset{
    basicstyle=\small\ttfamily,
    backgroundcolor=\color{gray!10}, 
    linewidth=\columnwidth,
    breaklines=true,
    frame=single, 
    rulecolor=\color{black}, 
    showstringspaces=false, 
    numbers=left, 
    numberstyle=\tiny\color{gray}, 
    xleftmargin=0em, 
    numbers=none, 
    moredelim=**[is][\color{gray!70}]{@}{@} 
}
\begin{lstlisting}[linewidth=\columnwidth,breaklines=true,caption={Vote - Verify Claim and Provide Verdict}, label={listing:verify_claim_bill}]
You are given the following claim, and 5 bills along with the bill title, a short summary of the bill, the vote cast on the bill and the alignment of the bill with the claim.
Alignment here means whether an individual bill supports, refines, or is irrelevant to the claim.
Your task is to give one of the given fact-check labels to the claim based on the evidence which are the bills.
You may consider factors such as the main objectives of the bill and unintended or implicit consequences.
Return your explanation and one of the following labels in JSON format.

Claim: {claim}

Bill Title 1: {bill_title_1}
Bill Summary 1: {bill_summary_1}
Vote 1: {vote_type_1}
Alignment1: {alignment_1}

[...]

Bill Title N: {bill_title_N}
Bill Summary N: {bill_summary_N}
Vote N: {vote_type_N}
Alignment N: {alignment_N}

Labels:
True: The given bills support the claim.
MostlyTrue: The given bills mostly support the claim.
HalfTrue: The given bills can only partly support or refute claim.
MostlyFalse: The given bills mostly refute the claim.
False: The given bills refute the claim.
Irrelevant: The given bills are not relevant to the claim.

Return the JSON object with the label and the explanation. The fields should be 'Label' and 'Explanation'.

\end{lstlisting}

\vspace{1cm}
\lstset{
    basicstyle=\small\ttfamily,
    backgroundcolor=\color{gray!10}, 
    linewidth=\columnwidth,
    breaklines=true,
    frame=single, 
    rulecolor=\color{black}, 
    showstringspaces=false, 
    numbers=left, 
    numberstyle=\tiny\color{gray}, 
    xleftmargin=0em, 
    numbers=none, 
    moredelim=**[is][\color{gray!70}]{@}{@} 
}
\begin{lstlisting}[linewidth=\columnwidth,breaklines=true,caption={OECD - Verify Claim and Provide Verdict}, label={listing:verify_claim_oecd}]
Your task is to verify the claim using the retrieved data and provide a verdict.
Claim: {claim}

Retrieved Data:
{formatted_data}

Instructions:
- Use only the provided data, regardless of its perceived relevance, to analyze the claim.
- The verdict must be based solely on the retrieved data. Do not rely on external knowledge.
- Ensure you consider the spirit of the claim in the fact-check and not just the precise verbage.

Verdict Categories:
- True: The statement is fully accurate, with no significant information missing.
- Mostly True: The statement is accurate but requires clarification or additional context.
- Half-True: The statement is partially accurate but omits important details or misrepresents the context.
- Mostly False: The statement contains some truth but overlooks key facts that would significantly alter the impression.
- False: The statement is completely inaccurate.

Output:
- Provide a verdict in the format:

Verdict: [False, Mostly False, Half-True, Mostly True, True]; Explanation: [Your reasoning].
\end{lstlisting}

\end{document}